\pdfoutput=1

\documentclass[11pt]{article}

\usepackage{acl}

\usepackage{times}
\usepackage{latexsym}
\usepackage{enumitem}
\usepackage{makecell}
\usepackage{graphbox}

\usepackage[T1]{fontenc}

\usepackage[utf8]{inputenc}

\usepackage{microtype}
\usepackage{array}
\usepackage{graphicx}
\usepackage{subcaption}
\usepackage{pgfplots}
\usetikzlibrary{patterns}
\usepackage{booktabs}
\usepackage{amsmath}
\usepackage{amssymb}
\usepackage{multirow,makecell}
\usepackage{lipsum}
\usepackage{color,soul}
\usepackage{pifont}
\usepackage{adjustbox}
\usepackage{amsfonts}
\usepackage{mathabx}

\usepackage{xcolor}
\usepackage{soul}

\newcommand{\hlc}[2][yellow]{{%
    \colorlet{foo}{#1}%
    \sethlcolor{foo}\hl{#2}}%
}

\definecolor{mygreen}{HTML}{00786C}
\definecolor{lightgrey}{HTML}{447777}
\definecolor{mypurple}{HTML}{D64C1D}
\definecolor{babyblue}{rgb}{0.54, 0.81, 0.94}
\definecolor{babypink}{rgb}{0.96, 0.76, 0.76}
\definecolor{brilliantlavender}{rgb}{0.96, 0.73, 1.0}

\newcommand{\ect}[1]{\textcolor{blue}{[#1]}}

\newcommand{\bvec}[1]{\boldsymbol{#1}}

\newcommand{\EG}[1]{\textcolor{mygreen}{#1}}
\newcommand{\CXT}[1]{\textcolor{mypurple}{#1}}
\newcommand{\RET}[1]{\textcolor{red}{#1}}

\newcommand{\model}[0]{\textsc{EgRet}}
\definecolor{mygray}{gray}{0.4}

\title{Modeling Exemplification in Long-form Question Answering via Retrieval}

\author{
{Shufan Wang}$^{1}$, \ {Fangyuan Xu}$^{2}$, {Laure Thompson}$^{1}$, \, {Eunsol Choi}$^{2}$, \ {Mohit Iyyer}$^{1}$ \\
$^{1}$ College of Information and Computer Sciences, University of Massachusetts Amherst\\
$^{2}$ Department of Computer Science, The University of Texas at Austin\\
{\tt \{shufanwang, laurejt, miyyer\}@cs.umass.edu } \\
{\tt \{fangyuan, eunsol\}@utexas.edu }
}

\begin{document}
\maketitle

\begin{abstract}
Exemplification is a process by which writers explain or clarify a concept by providing an example. While common in all forms of writing, exemplification is particularly useful in the task of long-form question answering (LFQA), where a complicated answer can be made more understandable through simple examples. In this paper, we provide the first computational study of exemplification in QA, performing a fine-grained annotation of different types of examples (e.g., hypotheticals, anecdotes) in three corpora. We show that not only do state-of-the-art LFQA models struggle to generate relevant examples, but also that standard evaluation metrics such as ROUGE are insufficient to judge exemplification quality. We propose to treat exemplification as a \emph{retrieval} problem in which a partially-written answer is used to query a large set of human-written examples extracted from a corpus. Our approach allows a reliable ranking-type automatic metrics that correlates well with human evaluation. A human evaluation shows that our model's retrieved examples are more relevant than examples generated from a state-of-the-art LFQA model.

\end{abstract}
\section{Introduction}
\label{sec:introduction}

When an author introduces a complicated concept, they commonly follow it up with a concrete example to help clarify their intended meaning. This process, known as  \emph{exemplification}, occurs in diverse forms, including hypothetical examples, personal anecdotes, and analogies~\citep{clouse2013student}. Exemplification is particularly common within the NLP task of \emph{long-form question answering} (LFQA), where an author wants to communicate a concept unknown to the question asker. 
Consider the following QA pair from the \texttt{r/explainlikeimfive} subreddit~\citep{fan-etal-2019-eli5}:

\begin{quote}
\small
\textbf{Q:} How does the ground not cave in while being under the heavy weight of cities?\\
\textbf{A:} \CXT{It's all about what they're building on, and occasionally they get it wrong...} \EG{For example, San Francisco's Millennium Tower, built on mud and sand, has already sunk 18 inches into the ground.}
\end{quote}

Here, the answerer uses a specific example to emphasize the importance of building on a solid foundation. In general, explaining via exemplification is a fundamental technique in these kinds of pedagogical scenarios~\citep{Hyland2007ApplyingAG}, and it warrants separate study due to its importance in LFQA and the challenges in evaluating it. 
However, existing work on building LFQA models~\cite{fan-etal-2019-eli5} does not give special treatment to exemplification or any other discourse phenomena, choosing instead to evaluate model outputs against reference answers using metrics such as ROUGE~\citep{lin2004rouge} that are not meaningful for this task~\citep{lfqa21}. In the above QA pair, any other structurally-unstable building (e.g., \EG{the Leaning Tower of Pisa}) could serve as a valid example, but an LFQA model would be unfairly penalized for generating one of these acceptable alternatives.

In this paper, we first conduct a detailed study of exemplification across three different domains: Wikipedia, web articles, and community answers to questions from the ELI5 LFQA dataset. We extract sentences and clauses associated with exemplification by matching on explicit markers such as \emph{``for example''} and \emph{``(e.g., ...)''} and annotate 300 examples from this dataset. Our analysis reveals significant variation in occurrence frequencies of different forms of exemplification (e.g., hypothetical vs. specific examples) across the three domains.


Next, we focus on improving the modeling and evaluation of the subtask of exemplification within LFQA. We propose to treat it as a \emph{retrieval} problem rather than a generation problem: given a question and a prefix of a reference answer (in the above QA pair, all of the text before \emph{``For example''}), a model  must retrieve the ground-truth example that follows the prefix (the sentence about the Millenium Tower) from the set of all exemplifying sentences and clauses in the dataset. We can use retrieval metrics such as recall@$k$ to evaluate a model's ability to select the ground-truth example, which are more informative than metrics such as ROUGE. 

We demonstrate that pretraining our retriever on a large-scale dataset of exemplifying units extracted from the Books3 corpus~\citep{pile} and then fine-tuning it on ELI5 examples results in substantial improvements on these ranking metrics. Finally, we crowdsource human evaluation comparing our retriever's outputs with those generated by the state-of-the-art ELI5 model of~\citet{lfqa21} and find that workers prefer the retriever's outputs far more frequently than those of the generation model. We hope that our work spurs more research into the evaluation and modeling of exemplification and other complex discourse phenomena present in LFQA. To facilitate future research, we publicly release the code and trained models from our work.\footnote{ \href{ https://github.com/north125ptlm/lfqa-retrieval }{ https://github.com/north125ptlm/lfqa-retrieval } }

\section{Exploring Exemplification}\label{sec:dataset}

In this section, we first describe our data extraction process, which we use to mine instances of exemplification from three datasets (and domains): ELI5~\citep{fan-etal-2019-eli5}, Natural Questions~\citep{natural_questions} and Books3~\cite{pile}. This process involves matching on a set of ``exemplification markers'' and collecting both the text of the matching example (a sentence or clause) as well as the surrounding context on both sides. We then conduct a fine-grained human annotation study on the extracted data, breaking exemplification down into different types and exploring how they are used across the different domains.

\subsection{Extracting a Dataset of Exemplification}

 \paragraph{Exemplification markers:} \citet{Hyland2007ApplyingAG} annotated a diverse collection of articles from  multiple disciplines with a variety of rhetorical practices\footnote{The annotation included texts from physics, biology, mechanical \& electric engineering, philosophy, sociology, applied linguistics, and marketing.} and found that more than 75\% of \EG{``examples"} are signalled parenthetically or lexically with the use of the three most frequent {``exemplification markers''}: \textit{``such as''}, \textit{``for example''}, \textit{``e.g.''}. Empirically, we find that the \textit{``such as''} marker is noisy at signalling exemplification and often leads to ambiguous cases where it is hard to automatically detect the example boundary. Hence, we take the other two most frequent {exemplification markers}, namely \textit{``for example''} and \textit{``e.g.''}, and extract the parentheses-enclosed clauses and sentences that contain these {exemplification markers} as \EG{examples}.
 
 
 \paragraph{Examples from Diverse Domains} Using these two exemplification markers, we extract a dataset of examples in context from two popular LFQA datasets that come from different domains, ELI5 \citep[][Reddit answers]{fan-etal-2019-eli5} and Natural Questions \citep[][Wikipedia passages]{natural_questions}.\footnote{We consider questions with \textit{only} a long answer span (i.e. paragraph answer) since they cannot be addressed solely by entity names or a boolean, and are suitable for studying LFQA.} To study the exemplification phenomenon from a more diverse perspective, we also extract \EG{examples} along with their surrounding \CXT{context} from the Books3 Corpus~\citep{pile}, a large-scale 100GB collection of books spanning a variety of topics and genres. Table \ref{tab:corpusstats} contains detailed statistics for the extracted example-in-context datasets.

\begin{table}
\small

\begin{center}
\begin{adjustbox}{max width=0.5\textwidth}

\begin{tabular}{ l r r r }  
 \toprule
  & ELI5 & NQ & Books3 \\
 \midrule
 \# training \EG{examples} & 65,157 & 1,209  & 2,848,171 \\
 \# validation \EG{examples} &  1,185 & 52  & 712,043 \\
 \midrule
 avg. \# \CXT{context} words & 123.7 & 74.0 & 155.7 \\
 avg. \# \EG{example} words & 23.3 & 33.1 & 27.1\\
 avg. \# right \CXT{words} tokens & 135.3 & 54.8 &  107.5\\
 \bottomrule
 
\end{tabular}
\end{adjustbox}

\end{center}
\caption{Statistics of extracted \EG{example}-in-\CXT{context} data.}
\label{tab:corpusstats}

\end{table}

\subsection{Fine-grained Annotation Study}
With the extracted dataset of examples, we conduct an in-depth analysis to understand different uses of exemplification in various domains. We (the authors) annotate a total of 300 examples extracted using exemplification markers from Natural Questions, ELI5 and Books3 as below. Fifty examples are annotated by two annotators (for purposes of computing agreement, reported in Section \ref{subsec:agreement}) and the rest are annotated by one annotator. 

Given an extracted example and its left and right context, we first filter out around 7\% of the extracted examples, either because they are extraction artifacts or because the marker is used for functions other than exemplification (e.g., referring to a figure or table). After this basic check, we annotate both structural information about the example (e.g., discourse units such as the anchor of the example) and semantic information about how it is used in the context. Table \ref{tab:annotation_stats} contains statistics of the annotated subset.\footnote{Annotated examples in the three datasets can be found in Table \ref{tab:annotated_example} in the appendix.} 

\begin{table}
\small

\begin{center}
\begin{adjustbox}{max width=0.5\textwidth}

\begin{tabular}{ l r r r }  
 \toprule
\textbf{Dataset} & \textbf{Valid} & \textbf{Extracted} & \textbf{\% Valid}  \\ 
 \midrule
 ELI5 & 87 & 93 &  94\% \\ 
 NQ &  89  & 95 &  94\% \\ 
 Books3 & 85 & 94 & 90\% \\ 
 \midrule
 Total & 261 & 282 & 93\% \\ 
 \bottomrule
 
\end{tabular}
\end{adjustbox}

\end{center}\vspace{-0.4em}
\caption{Statistics of annotated \EG{examples}.}
\label{tab:annotation_stats}

\end{table}

\subsubsection{Discourse units} Exemplification is usually expressed through three discourse units~\cite{Meyer1992AppositionIC, Triki2021ExemplificationIR}: the anchor (also known as ``exemplified unit''), the exemplification marker, and \EG{the example text} itself (``exemplifying unit''). We annotate the anchor (marked as \textbf{bold}) and example (marked as \textit{italics}). Concretely, in example (1) below, the anchor is ``\CXT{euryhaline species}'', the exemplifying marker is ``e.g.'', and the exemplifying unit is ``\EG{Flounder}''. 
As in the study of~\citet{Triki2021ExemplificationIR}, we find that these units mainly come in two forms: (1) nominal groups that refer to entities,  or (2) clauses that represent statements.
\begin{quote}
\small
(1) \CXT{However , some fish show a tremendous ability to effectively osmoregulate across a broad range of salinities ; fish with this ability are known as \textbf{euryhaline species}} , e.g., \EG{\textit{Flounder}}.
\end{quote}

\begin{quote}
\small
(2) \CXT{\textbf{Players earn you points, depending on their performance.}} For example, \EG{\textit{your QB might throw for 200 yards, earning 1 point per 10 yards, for 20 points.}}
\end{quote}

During our initial investigation, we also noticed examples (3) which are signalled \textit{implicitly} (i.e. without exemplification markers).\footnote{Recent study on long form QA~\cite{xu2022lfqadiscourse} contains manually identified example sentences, including those signalled implicitly.} Identifying such examples automatically is beyond the scope of our study and warrants future work.
\begin{quote}
\small
(3) \CXT{\textbf{The biggest driver of disparity in tech jobs is cost of living.}} \EG{\textit{If it costs 2000 a month to live in Boston, and 200 a month to live in India, then salaries will reflect that.}}
\end{quote}

Table \ref{tab:clause_length} shows that the length of the discourse units is roughly the same across the three datasets, which supports our experimental decision in Section \ref{sec:experiments} of using sentences as the base unit for example retrieval. We also find that all of the anchors we annotated occur \textit{before} the examples, suggesting that using preceding \CXT{context} to retrieve examples gives the model sufficient information.

\begin{table}
\small

\begin{center}
\begin{adjustbox}{max width=0.5\textwidth}

\begin{tabular}{ l r r r }  
 \toprule
 \textbf{Dataset} & \textbf{\# samples} & \textbf{Anchor} & \textbf{Example}  \\
 \midrule
ELI5 & 87 & 1.1/16.6 & 1.3/29.2 \\
NQ & 89 & 1.0/15.4 & 1.4/25.1 \\
Books3 & 85 & 1.1/17.0 & 1.2/25.1 \\
 \midrule
 \textbf{Type} \\ 
 \midrule
 Real & 209 & 1.1/14.5 & 1.3/24.6 \\
 Hypothetical & 52 & 1.2/23.4 & 1.2/33.7 \\  
 \midrule
 Personal & 13 &  1.1/18.1 & 2/49.6 \\
 Not-personal & 248 & 1.1/16.3 & 1.2/25.2 \\
 \bottomrule
 
\end{tabular}
\end{adjustbox}

\end{center}
\caption{Length of the discourse units per dataset and type, presented as average \# sentences / \# words.}
\label{tab:clause_length}

\end{table}

\subsubsection{Real vs. Hypothetical Examples}
One notable categorization found during our investigation and also identified by~\citet{Triki2021ExemplificationIR} is whether the examples are real, specific scenarios, or hypothetically-constructed scenarios. We detail the definition of the two types here:

\paragraph{Real examples:} These examples are either real entities (4) or specific scenarios (5) that are constructed as fact clauses. 

\begin{quote}
\small
(4) \CXT{CEOs lead a range of organizations , including public and private corporations, non-profit organizations and even \textbf{some government organizations}} (e.g., \EG{\textit{Crown corporations}}).
\end{quote}

\begin{quote}
\small
(5) \CXT{\textbf{For a given pressure , different liquids boil at different temperatures}.} For example, \EG{\textit{water boils at 100° C (212° F) at sea level, but at 93.4° C (200.1° F) at 2,000 metres (6,600 ft) altitude}.}
\end{quote}

\paragraph{Hypothetical examples:} In contrast, hypothetical examples are scenarios constructed by the author. According to \citet{Triki2021ExemplificationIR}, hypothetical examples often come with the use of conditional clauses \textit{if} or signalled via \textit{assume}.  These examples are generally more complicated and are specifically constructed for the purpose of exemplification.

\begin{quote}
\small
(6) \CXT{\textbf{The reasoning is that if you share your life with someone then anything you do during that time is made possible by their support.}} For example, \EG{\textit{if your wife is a stay at home wife and your a business man making lots of money, the reasoning is that you would not have the same amount of time to dedicate to work if you had to look after your own house and/or children.}}
\end{quote}

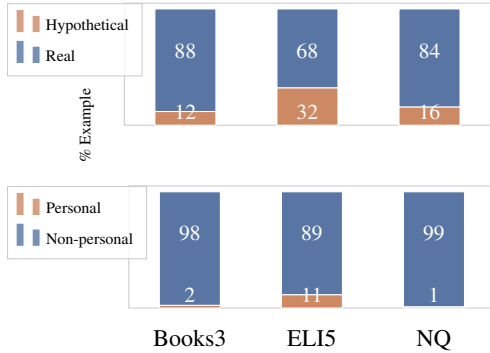
\begin{figure}
\begin{subfigure}{1\textwidth}
\begin{tikzpicture}

\definecolor{color1}{rgb}{0.347058823529412,0.458823529411765,0.641176470588235}
\definecolor{color0}{rgb}{0.798529411764706,0.536764705882353,0.389705882352941}

\begin{axis}[
font=\small,
axis line style={white!80!black},
legend cell align={left},
legend style={fill opacity=0.8, 
draw opacity=1, text opacity=1, at={(0.05, 1)}, draw=white!80!black, font=\tiny},
height=.2\textwidth,
width=.4\textwidth,
tick align=outside,
x grid style={white!80!black},
xmin=-0.5, xmax=2.5,
xtick style={color=white},
xtick={0,1,2},
xticklabel,
y grid style={white!80!black},
ylabel={\% Example},
y label style={at={(axis description cs:.2,0)},anchor=south, font=\tiny},
ymajorticks=false,
ymin=0, ymax=105,
ytick style={color=white!15!black}
]
\draw[draw=white,fill=color0] (axis cs:-0.25,0) rectangle (axis cs:0.25,11.7647058823529);
\addlegendimage{ybar,ybar legend,draw=white,fill=color0}
\addlegendentry{Hypothetical}

\draw[draw=white,fill=color0] (axis cs:0.75,0) rectangle (axis cs:1.25,32.183908045977);
\draw[draw=white,fill=color0] (axis cs:1.75,0) rectangle (axis cs:2.25,15.7303370786517);
\draw[draw=white,fill=color1] (axis cs:-0.25,11.7647058823529) rectangle (axis cs:0.25,100);
\addlegendimage{ybar,ybar legend,draw=white,fill=color1}
\addlegendentry{Real}

\draw[draw=white,fill=color1] (axis cs:0.75,32.183908045977) rectangle (axis cs:1.25,100);
\draw[draw=white,fill=color1] (axis cs:1.75,15.7303370786517) rectangle (axis cs:2.25,100);

\draw (axis cs:0,-2.5) node[
  scale=0.9,
  anchor=south,
  text=white,
  rotate=0.0
]{12};
\draw (axis cs:0,50) node[
  scale=0.9,
  anchor=south,
  text=white,
  rotate=0.0
]{88};

\draw (axis cs:1,-2.5) node[
  scale=0.9,
  anchor=south,
  text=white,
  rotate=0.0
]{32};

\draw (axis cs:1,50) node[
  scale=0.9,
  anchor=south,
  text=white,
  rotate=0.0
]{68};

\draw (axis cs:2,-2.5) node[
  scale=0.9,
  anchor=south,
  text=white,
  rotate=0.0
]{16};

\draw (axis cs:2,50) node[
  scale=0.9,
  anchor=south,
  text=white,
  rotate=0.0
]{84};
\end{axis}

\end{tikzpicture}
\end{subfigure}
\begin{subfigure}{1\textwidth}
\begin{tikzpicture}

\definecolor{color1}{rgb}{0.347058823529412,0.458823529411765,0.641176470588235}
\definecolor{color0}{rgb}{0.798529411764706,0.536764705882353,0.389705882352941}

\begin{axis}[
font=\small,
axis line style={white!80!black},
legend cell align={left},
legend style={fill opacity=0.8, 
draw opacity=1, text opacity=1, at={(0.05, 1)}, draw=white!80!black, font=\tiny},
height=.2\textwidth,
width=.4\textwidth,
tick align=outside,
x grid style={white!80!black},
xmin=-0.5, xmax=2.5,
xtick style={color=white},
xtick={0,1,2},
xticklabel,
xticklabels={Books3,ELI5,NQ},
y grid style={white!80!black},
y label style={at={(axis description cs:.2,0)},anchor=south, font=\tiny},
ymajorticks=false,
ymin=0, ymax=105,
ytick style={color=white!15!black}
]
\draw[draw=white,fill=color0] (axis cs:-0.25,0) rectangle (axis cs:0.25,2.35294117647059);
\draw[draw=white,fill=color1] (axis cs:-0.25,2.35294117647059) rectangle (axis cs:0.25,100);

\addlegendimage{ybar,ybar legend,draw=white,fill=color0}
\addlegendentry{Personal}

\draw[draw=white,fill=color0] (axis cs:0.75,0) rectangle (axis cs:1.25,11.4942528735632);
\draw[draw=white,fill=color1] (axis cs:0.75,11.4942528735632) rectangle (axis cs:1.25,100);

\draw[draw=white,fill=color0] (axis cs:1.75,0) rectangle (axis cs:2.25,1.12359550561798);
\draw[draw=white,fill=color1] (axis cs:1.75,1.12359550561798) rectangle (axis cs:2.25,100);

\addlegendimage{ybar,ybar legend,draw=white,fill=color1}
\addlegendentry{Non-personal}

\draw (axis cs:0,-2.5) node[
  scale=0.9,
  anchor=south,
  text=white,
  rotate=0.0
]{2};
\draw (axis cs:0,50) node[
  scale=0.9,
  anchor=south,
  text=white,
  rotate=0.0
]{98};

\draw (axis cs:1,-2.5) node[
  scale=0.9,
  anchor=south,
  text=white,
  rotate=0.0
]{11};

\draw (axis cs:1,50) node[
  scale=0.9,
  anchor=south,
  text=white,
  rotate=0.0
]{89};

\draw (axis cs:2,-2.5) node[
  scale=0.9,
  anchor=south,
  text=white,
  rotate=0.0
]{1};

\draw (axis cs:2,50) node[
  scale=0.9,
  anchor=south,
  text=white,
  rotate=0.0
]{99};

\end{axis}

\end{tikzpicture}
\end{subfigure}
\caption{Distribution of different types of examples across three datasets.}
\label{fig:real_hypothetical}
\end{figure}

We observe a different distribution of the two types of examples in the three datasets (Figure \ref{fig:real_hypothetical}, top). ELI5 contains more hypothetical examples (32\%) than the other two datasets (16\% for NQ, 12\% for Books3), showing that hypothetical examples are commonly used to explain complicated concepts. 
We also note that hypothetical examples are generally longer than real ones (33.7 v.s. 24.6 words, as seen in Table \ref{tab:clause_length}), aligning with our observation that these examples are more complicated.

\subsubsection{Personal Information}
Previous work found that exemplification is ``rarely personalized'' in academic texts~\cite{del2006MetadiscourseIL}, referring to the uncommon use of first person pronouns or reference to the author. In contrast, we found a consistent presence of personal information in the examples we examined, in the form of either personal anecdotes (7) or an example situated in the author's own circumstance (e.g., \EG{\textit{in my city...}}). We thus also annotate whether the exemplifying units contain personal information or not.

\begin{quote}
\small
(7) \CXT{... \textbf{But they also give advice a doctor might have forgotten.}} For example \EG{\textit{about 6 months ago I went to the urgent care center for ear pain and was prescribed an ear drop antibiotic.}} ...
\end{quote}

We observe differences in the presence of personal information across the three datasets (Figure \ref{fig:real_hypothetical}, bottom) -- ELI5 answers, which are written by users from online community forum, contain a substantial portion of examples with personal information (12\%), while such information is rarely present in the other two datasets. There is also a notable length difference between examples with and without personal information (49.6 v.s. 25.2 words), showing that more detailed description is provided for personal anecdotes. 
The observation that ELI5 contains many personal examples raises concerns that language models trained on such datasets will generate personal examples that cannot be verified or meaningfully interpreted.
\subsection{Annotation agreement}\label{subsec:agreement}
We report agreement for the three annotation tasks performed on the 50 two-way annotated samples. For discourse unit annotation, we find high unigram overlap between the two annotations (0.81 for  the anchor and 0.92 for the example). For annotation of real vs. hypothetical examples and the presence of personal information, we find a modest to high agreement with a Cohen's kappa of 0.48 for both. Additionally, we observe annotations of real v.s. hypothetical example to be split when the example refers to abstract actions, such as the one below (8).

\begin{quote}
\small
(8) Carpets are used for a variety of purposes , including insulating a person 's feet from a cold tile or concrete floor , \CXT{\textbf{making a room more comfortable}} as a place to sit on the floor (\textit{e.g. , \EG{when playing with children or as a prayer rug }}) , reducing sound from walking ( particularly in apartment buildings ) and adding decoration or colour to a room .
\end{quote}

\section{Retrieving Examples in LFQA}
\label{sec:experiments}

\begin{figure*}[ht]
\centering
\includegraphics[width=\linewidth]{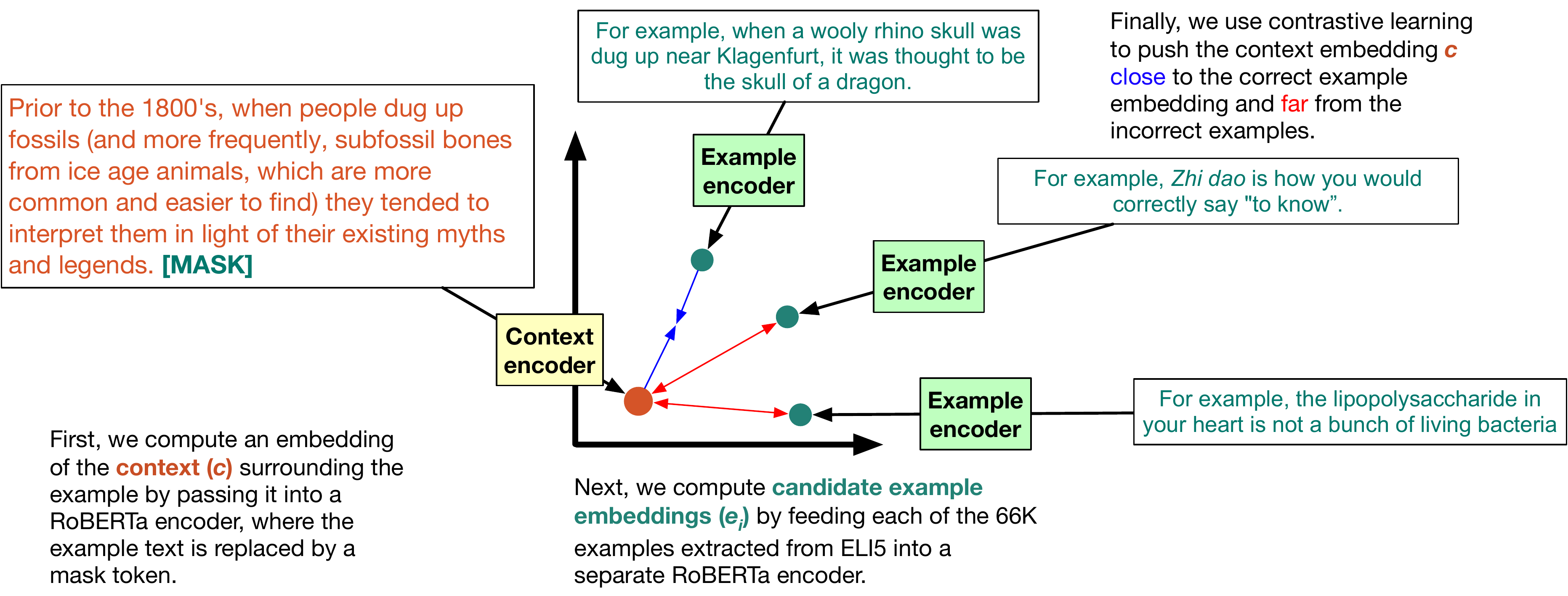}
\caption{Our \model\ model uses dual RoBERTa encoders to embed (1) the context surrounding an exemplifying unit and (2) all 66K exemplifying units extracted from ELI5. A contrastive objective is then used to move the context embedding close to the embedding of the ground-truth exemplifying unit that occurs within that context, and far from all other negative exemplifying units.}
\label{fig:model_example}
\end{figure*}

Our annotation and analysis of the extracted exemplification dataset reveals the diversity and complexity of this discourse phenomenon. In this section, we shift our focus towards building retrieval-based models to produce \EG{examples} based on a given LFQA \CXT{context}. First, we define our \EG{example}-retrieval task formally and introduce our evaluation metrics. Then, we describe our contrastive-learning based retriever model, which closely follows the retriever implementation in \citet{thairelic22}, and baseline retrieval models, and report their performances. 

\subsection{Task Definition}
Given a \CXT{context} (part of the answer to a given LFQA question) with a masked out \EG{exemplifying unit}, model is asked to retrieve that masked unit from a retrieval corpus. We consider two settings for \CXT{context}: (1) concatenation of left and right contexts surrounding the \EG{exemplifying unit} and (2) left context preceding the \EG{exemplifying unit} only (as in Figure~\ref{fig:model_example}). Both left and right contexts are truncated to 256 tokens surrounding the exemplifying unit. We use all 66K exemplifying units extracted from the 272K QA instances in the training and development portion of ELI5 dataset~\cite{petroni-etal-2021-kilt} as the retrieval corpus. 
To illustrate the input format, consider the below answer to the question ``Why didn't anyone discover dinosaur fossils before the 1800s?'':
\begin{quote}
    \small
    \CXT{Prior to the 1800's, when people dug up fossils (and more frequently, subfossil bones from ice age animals, which are more common and easier to find) they tended to interpret them in light of their existing myths and legends.} \EG{\bf \textsc{[mask]}} \CXT{Because fossils are almost always found as a jumble of bones rather than a neat skeleton and because they are incomplete... nobody looked at dinosaur skeletons and realized what the animals that made them actually looked like.}
\end{quote}

where the \EG{\bf \textsc{[mask]}} token corresponds to 

\begin{quote}
    \small
    \EG{For example, when a wooly rhino skull was dug up near Klagenfurt, it was thought to be the skull of a dragon.}
\end{quote}

 The first quote block showing \CXT{the masked answer} will be used as a query to retrieve \EG{the exemplifying unit} in the second quote block.

This retrieval task is challenging due not only to the size of the candidate set but also because of the topical similarity between many ELI5 questions, which was previously noted by~\citet{lfqa21}. Retrieving based on lexical overlap, as in BM-25 and other string-matching based approaches, cannot identify exemplifying units that are relevant but share little overlap with the context.

\subsection{Evaluation Data / Metric}

For evaluation data, we use 1,185 \CXT{context}-\EG{example} pairs automatically identified from 1,507 QA instances in the ELI5 development set~\cite{petroni-etal-2021-kilt} by our discourse marker heuristics.  
We evaluate the performance of our retriever by measuring how reliable it is at retrieving ground-truth examples from the candidate example set. Concretely, given the candidate set of \EG{examples}, each model should output a ranked list of all \EG{examples} according to their fit for the \CXT{context}. We evaluate these rankings using recall@$k$  of the ground-truth \EG{example} (where  $k=1, 3, 5, 10, 50, 100$) from the set of all 66K examples in ELI5. By using retriever-based evaluation metrics, we can directly measure a model's ability to understand and exemplify a context, in contrast to string overlap metrics like ROUGE~\citep{lin2004rouge} which are uninformative for this task.

\begin{table*}[ht]
\small
\begin{center}
\begin{tabular}{ lrrrrrrrrr } 
 \toprule
 \bf Model & \bf Context & \multicolumn{6}{c}{\bf Recall@\emph{k} ($\uparrow$)} & \bf Avg rank ($\downarrow$)  \\
\cmidrule{3-8}\vspace{-0.3cm}\\
  & & 1 & 3 & 5 & 10 & 50 & 100 &    \\
 \midrule
 \multicolumn{9}{c}{\textit{\ect{zero-shot} baselines, including pretrained dense retrievers}}\vspace{0.1cm}\\
 Random & & 0.0 & 0.0 & 0.01 & 0.02 & 0.08 & 0.15 & 33171.0 \vspace{0.1cm}\\
 BM25~\citep{robertson1995okapi} & L & 4.6 & 9.5 & 12.1 & 16.2 & 25.6 & 30.4 & 24940.1 \\
 DPR~\citep{karpukhin-etal-2020-dense} & L & 2.7& 5.2 & 7.1 & 9.7 & 20.3 & 27.5 & 7796.9 \\
 ColBERT~\citep{khattab2020colbert} & L & 6.0 & 11.8 & 14.3 & 18.2 & 31.2 & 36.3 & 9948.6 \\
 SBERT~\citep{reimers-2020-Curse_Dense_Retrieval} & L & 5.7 & 11.6 & 15.0 & 20.4 & 34.3 & 42.2 & 5122.7\vspace{0.15cm} \\
 
 BM25~\citep{robertson1995okapi} & L+R & 8.7 & 16.1 & 20.0 & 24.8 & 38.0 & 42.7 & 13968.8 \\
 DPR~\citep{karpukhin-etal-2020-dense} & L+R & 4.4 & 8.6 & 11.1 & 15.7 & 27.2 & 33.5 & 5684.0\\ 
 ColBERT~\citep{khattab2020colbert} & L+R & 8.9 & \textbf{16.4} & 18.9 & 23.3 & 36.3 & 42.2 & 8636.0 \\
 SBERT~\citep{reimers-2020-Curse_Dense_Retrieval} & L+R & \textbf{9.2} & 16.3 &\textbf{ 21.2} & \textbf{27.1} & \textbf{44.2} & \textbf{51.3} & \textbf{3216.7}\\
\midrule  
\multicolumn{9}{c}{\textit{our models, trained on exemplification data}}\vspace{0.1cm}\\
 \model\ (ELI5) & L & 13.0 & 22.8 & 29.3 & 36.5 & 55.2 & 64.0 & 807.2 \\
 \model\ (Books3 only) & L & 19.3 & 30.4 & 36.8 & 44.1 & 63.1 & 69.0 & 2427.6 \\  
 \model\ (Books3 + ELI5) & L & 21.1 & 33.5 & 39.2 & 46.8 & 66.7 & 73.0 & 609.8\vspace{0.15cm}\\ 
 \model\ (ELI5)  & L+R & 23.5 & 35.6 & 41.9 & 51.0 & 71.0 & 77.6 & 300.6 \\
 \model\ (Books3 only) & L+R & 33.4 & 47.8 & 53.8 & 61.5 & 76.5 & 82.5 & 632.9\\  
 \model\ (Books3 + ELI5) & L+R & \textbf{36.9} & \textbf{50.6} & \textbf{58.2} & \textbf{64.8} &\textbf{ 80.2} & \textbf{85.8} & \textbf{188.4}  \\
\bottomrule
\end{tabular}
\end{center}
\caption{Our \model\ model outperforms pretrained (or non-parametric) baselines  on the example retrieval task, indicating that exemplification cannot be solved by term matching or coarse query-context similarity alone.  Pretraining \model\ on out-of-distribution examples from Books3 results in large improvements in recall@$k$. Finally, including \CXT{context} to the right of the \EG{exemplifying unit} significantly boosts performance.}
\label{tab:retriever_stats}
\end{table*}

\subsection{Models}
We first introduce our model (\model) and describe baseline retrieval methods. We use all baseline models in a zero-shot manner (without additional fine-tuning on exemplification dataset).


\paragraph{\model: an example retriever for LFQA}
We train an example retriever model (\model) on our extracted exemplification dataset from training portion of ELI5 \citep{fan-etal-2019-eli5}, which contains $65K$ extracted \CXT{context}-\EG{example} pairs.
Our retriever consists of dual Transformer encoders~\cite{Vaswani2017AttentionIA}, one to encode the context query and the other to encode candidate examples (Figure~\ref{fig:model_example}). Both encoders are initialized as pretrained RoBERTa-base models~\cite{liu2019roberta}, as in the dense-RELiC model of~\citet{thairelic22}. To obtain a query embedding \CXT{$\bvec{c}_i$}, we feed the query encoder the surrounding \CXT{context} of a masked-out \EG{example}, as in Figure~\ref{fig:model_example}. Similarly, we use the other encoder to compute embeddings of the ground-truth example (\EG{$\bvec{e}_i^+$}) as well as negative examples sampled from other contexts, forming a set $E$ of example embeddings. We fine-tune both encoders in \model\ with a contrastive learning objective~\cite{infoNCE,chen2020simple}:
\begin{align}
    \mathcal{L}(\theta) &= - \sum_{(\CXT{\bvec{c_i}}, \EG{\bvec{e_i}}) \in E} \log \frac{\exp \CXT{\bvec{c}_i} \cdot \EG{\bvec{e_i^+}} }{\sum_{\EG{\bvec{e_j}} \in E} \exp \CXT{ \bvec{c}_i} \cdot \EG{\bvec{e_j}} }
\end{align}
This objective places the context vector $\CXT{\bvec{c_i}}$ \textit{close} to that of the ground-truth example vector $\EG{\bvec{e_i^+}}$ of an example, and \textit{far} from other examples 
$\EG{\bvec{e_j}}$ in the batch $E$ (``in-batch'' negative samples). 
We train both the left-context-only and the left-and-right-context models on a single RTX-8000 GPU for 10 epochs, using the Adam optimizer \citep{adam_kingma} with learning rate initialized at $1e-5$ for 10 epochs with early stopping. Both models converge in 4 epochs of training over the ELI5 dataset.

\paragraph{Pretraining \model\ on a huge set of examples:}
While the \model\ model described above is trained on the ELI5 dataset, exemplification is pervasive in many kinds of texts, as shown by our annotation in Section~\ref{sec:dataset}. Thus, we also experiment with a \emph{transfer} learning scenario by pretraining \model\ on a dataset of 3.5 million examples extracted from Books3 \citep{pile}, and then fine-tuning the resulting model on the ELI5 examples. We perform Books3 pretraining for both left-context-only and  left-and-right-context models on a single RTX-8000 GPU for 5 epochs using Adam with learning rate initialized at $1e-5$. Both models converge after one epoch of fine-tuning over the ELI5 dataset.

\paragraph{Baselines:} 
We compare \model\ to a term matching method as well as three publicly-available pretrained dense retrievers.

\noindent\textbf{BM25~\citep{robertson1995okapi}:}  BM25 retrieves text via a scoring function reliant on lexical overlap. We use the implementation from the rank\_bm25 library,\footnote{https://github.com/dorianbrown/rank\_bm25} with the default BM25Okapi as the similarity algorithm. 

\noindent\textbf{\textbf{DPR}~\citep{karpukhin-etal-2020-dense}:} DPR is a retriever model trained on Natural Questions~\citep{natural_questions} that computes dense representations of queries and evidence paragraphs for retrieval. 

\noindent\textbf{ColBERT~\citep{khattab2020colbert}:} ColBERT similarly uses pretrained language models to embed text, but contextualizes query and candidate documents using late interaction and was trained on MS MARCO~\citep{bajaj2018ms}.

\noindent\textbf{SBERT~\citep{reimers-2020-Curse_Dense_Retrieval}:} SBERT is a {sentence-BERT}-based encoder model with down-project layers that outperforms DPR on the MS MARCO dataset.

\section{Results \& Analysis}
\label{sec:result}

We report results from baseline retrievers and our trained models in Table~\ref{tab:retriever_stats}. We also conduct a human evaluation to compare retrieved examples from \model\ (\textbf{L}) to examples \emph{generated} by the state-of-the-art LFQA Routing Transformer model of~\citet{lfqa21}.

\subsection{Automatic Evaluation}
 While all models outperform random baselines, our trained \model\ models \citep{fan-etal-2019-eli5} consistently outperform both the lexical baseline (BM25) and other neural baselines. Among the baseline models, SBERT model, trained on MSMARCO dataset, consistently outperforms other models and DPR model lags behind the lexical baseline.\footnote{Models that outperform others in recall@$k$ when $k$ is small (less than 100) but underperform others when $k$ is large (greater than 1000) typically score poorly in the average rank (e.g. ColBERT). More details about the variation in these retrieval evaluation metrics can be found in the Appendix \S \ref{appendix:recall@k}.}  
\paragraph{Including \CXT{context} after the \EG{exemplifying unit} improves recall:} As a sanity check, we observe that including more context (\textbf{L+R}) significantly boosts recall for both \model\ and all baselines compared to including just context before the exemplifying unit (\textbf{L}). In addition to providing more constraints over the exemplifying unit, we also observe improvements on multi-sentence exemplifying units due to term matching; as this \textbf{L+R} setting is not particularly realistic, we analyze only the \textbf{L} configuration moving forward.

\begin{table*}[ht]
\scriptsize

\begin{center}

\begin{tabular}{m{3cm}m{3cm}m{5.5cm}m{3cm}}  
 \toprule
 \textbf{Left context}  & \textbf{Ground-truth} & \textbf{\hlc[babyblue]{\textit{EGRET}}-retrieved} vs \textbf{others} & \textbf{Analysis} \\

  \midrule
  \CXT{Evolution is not a force towards the optimum, it's a force towards the minimum necessary. }
  
  & \EG{For example, if grass was poisonous, it would be better for its survival, as less animals would come eat it.}  
  
  &\textbf{[\hlc[babyblue]{\textit{EGRET}}-retrieved}]: For example, we move incredible slowly when compared to the maximum speed allowed in the universe. (\textbf{R}:0.111 / \textbf{H}:5.0) 
  
  \vspace{0.15cm}
  
   \textbf{[Highest ROUGE-L]:} For example, if a certain percentage of snakes are venomous, then the more snakes in a given area the more venomous. (\textbf{R}:0.22)  

  & ROUGE is not a viable evaluation for example quality. Our \hlc[babyblue]{\textit{EGRET}}'s retrieved example was rated a 5/5 by all three crowdworkers but achieves lower ROUGE than an irrelevant example. \\

  \midrule
  \CXT{... You're brain is asleep and not paying any attention to your body so it ignores all of these stimuli unless they become too hard to ignore.} 
  
  & \EG{For example if the touching turns to slapping, the talking turns to yelling, or the light in the eyes turns to really bright light in the eyes.}

  & \textbf{[\hlc[babyblue]{\textit{EGRET}}-retrieved]}: For example, if you're in a room with a clock ticking you don't notice the ticking after a while. (\textbf{H}:4.0)
  
  \vspace{0.1cm}
  
  \textbf{[\hlc[brilliantlavender]{\textit{c-REALM-RT}} Generated]}: For example, its not just that your brain is dead. (\textbf{H}:2.0)
  
  & The \textbf{\hlc[babyblue]{\textit{EGRET}}-retrieved} example effectively illustrates the phenomenon in the context and receives a higher average rating  from crowdworkers than the generated example and even the ground-truth (3.3).  \\

  \midrule
  \CXT{... Multiple births mean less time per offspring. Each individual offspring therefore has a lower chance of survival, ... Seems like the larger mammals tend to have \textbf{single births.}}

  & \EG{For example, polar bears and elephants usually have single births.}
  
  & \textbf{[\hlc[babyblue]{\textit{EGRET}}-retrieved]}: \RET{For example, in mammals, a typical litter will be one offspring per pair of nipples as this is as many individuals a female can reasonably sustain.}
  \vspace{0.1cm}
  \textbf{[\hlc[brilliantlavender]{\textit{c-REALM-RT}} Generated]}: \textcolor{red}{For example, ok, this is as close I can get to explaining in ELI5 terms. }
  
  & \hlc[babyblue]{\textit{EGRET}}\ retrieves an example based on a key entity from the context (mammals) but fails to address the concept to be exemplified (\textbf{``single births''}) 
  \\

 \bottomrule
 
\end{tabular}

\end{center}
\caption{Instances where \hlc[babyblue]{\textit{EGRET}}\ retrieves exemplifying units that are rated highly by \textbf{H}umans but have low \textbf{R}OUGE score with the ground-truth example (top);  where model retrievals are rated as more meaningful than those generated by the \hlc[brilliantlavender]{\textit{c-REALM-RT}} model (middle); and where \hlc[babyblue]{\textit{EGRET}}\ \RET{fails} to retrieve a relevant example by relying too much on lexical overlap and \hlc[brilliantlavender]{\textit{c-REALM-RT}} \RET{fails} by producing an overly-generic output  (bottom). }
\label{tab:analysis}

\end{table*}

\paragraph{Pretraining improves example retrieval:}
Pretraining on out-of-distribution Books3 examples substantially boosts \model's performance, with the best left-context-only model achieving a recall@1 of 21.1\% compared to 13.0\% without pretraining. In fact, an \model\ model pretrained on Books3 \emph{without} fine-tuning on in-domain ELI5 data (19.3\% recall@1) outperforms the ELI5-only \model. While Figure~\ref{fig:real_hypothetical} shows that the distribution over exemplification types differs based on the dataset/domain, our results suggest that many aspects of exemplification apply generally to wide forms of writing, and that the pretrained Books3 \model\ could be useful for many other applications.

\subsection{Human evaluation of retrieved examples vs. generated examples}
How do the examples retrieved by \model\ compare to examples generated by a state-of-the-art LFQA model?  In theory, generative LFQA models can produce examples tailored to any input context; in practice, however, they struggle to generate relevant and informative examples. On the other hand, retriever models will always produce human-written examples, but it may not always be possible to retrieve a relevant example for an arbitrary held-out context. To explore this trade-off, we conduct a human evaluation by providing Mechanical Turk workers with an ELI5 question and \CXT{context}, and asking them to both \emph{rank} and \emph{rate} (on a 5 point Likert scale) three candidate \EG{exemplifying units}: (1) the ground-truth; (2) the top-ranked retrieval from \model, restricted to only cases where this retrieval is \emph{not} the ground-truth; and (3) a generated output from the state-of-the-art c-REALM-RT model of~\citet{lfqa21}.\footnote{This model is pretrained on the PG-19 dataset~\citep{Rae2020Compressive} and fine-tuned on ELI5, conditioned on retrieved Wikipedia documents.}

\paragraph{Task setup:}
In the ranking task, we ask workers to produce a ranking of the three choices (e.g., 1>2>3). We allow equality (e.g., 1=2>3) since multiple candidates can be equally valid for a given context. In the rating task, we ask workers to evaluate how well each \EG{example} fits with the given \CXT{context} on a scale of 1 to 5. For both tasks, we collect three annotations per item for 100 total items, and we pay \$0.35 per item for an estimated hourly wage of \$21 per hour.\footnote{We restrict workers to those in English speaking countries who have completed at least 1000 HITs with an acceptance rate of 97\%.} While a completely fair comparison of \model\ to c-REALM-RT is infeasible due to differences in training objective and architecture, we choose to focus only on sentence-level \EG{exemplifying units} that begin with \textit{``For example''}. We provide the question, left \CXT{context}, and \textit{``For example''} marker to c-REALM-RT and decode using nucleus sampling with $p=0.9$ until a sentence boundary (e.g., period) is generated. For the retrieved output, we use \model (Books3 + ELI5, \textbf{L}) since the RT model has access to only the left context.

\begin{table}[t]
    \centering\small
        \begin{adjustbox}{max width=0.48\textwidth}

\begin{tabular}{lrrc}
\toprule

  \bf  & $\textbf{Rating}_{\text{STD}}$ ($\uparrow$) & \bf Krip. $\alpha$  \\
  \midrule
  c-REALM-RT & $2.80_{0.775 }$ & 0.058 \\ 
  
  \model\ (Books3 + ELI5, \textbf{L}) & $3.55_{0.636}$ & 0.125 \\ 
  
  Ground-truth & $3.70_{0.597}$ & 0.128 \\
  
  \bottomrule
\end{tabular}
\end{adjustbox}
    \caption{Crowdworkers rate \model\ retrievals higher (on a scale of 1 to 5 of how well the \EG{exemplifying unit} fits with the \CXT{context}) than the SOTA generative LFQA model.}
    \label{tab:human_rating}
\end{table}

\begin{table}[t]
    \centering\small
    \begin{adjustbox}{max width=0.48\textwidth}

\begin{tabular}{lrrc}
\toprule

  \bf  & \bf $\textbf{Ranking}_{\text{STD}}$ ($\downarrow$) & \bf Krip. $\alpha$  \\
  \midrule
  c-REALM-RT & $2.26_{0.271}$ & 0.168 \\ 
  
  \model\ (Books3 + ELI5, \textbf{L}) & $1.88_{0.252}$ & 0.154 \\ 
  
  Ground-truth & $1.71_{0.284}$ & 0.200 \\
  
  \bottomrule
\end{tabular}

    \end{adjustbox}
    \caption{On our ranking task (1=best, 3=worst), crowdworkers prefer \model\ retrievals over the generative LFQA model.}
    \label{tab:human_ranking}
\end{table}


\paragraph{\model\ retrievals are preferred over generated \EG{exemplifying units}:}

In both tasks, crowdworkers exhibit a clear preference for \EG{exemplifying units} retrieved by \model\ compared to those generated by c-REALM-RT. While both tasks are fairly subjective, as shown by the low inter-annotator agreement measured by Krippendorf's alpha, {the results indicates} 
that as of now, exemplification in LFQA is better handled by retrieval models than generative models, and that research into hybrid generation/retrieval LFQA models is a promising direction. 

One limitation of our retrieval approach is that it will fail when the  candidate set contains no relevant examples for a given context, which is not (at least in theory) an issue with generative models.  However, we observe from our error analysis (Table \ref{tab:example_error_analysis}) that both approaches at times produce seemingly relevant but incorrect examples. Furthermore, our human evaluations show that generative model fails to produce relevant examples for 79\% of the contexts and consistently receive lower ratings than the retrieval model. This gap indicates that effectively incorporating retrieved information into the answer generation process is an important future research direction.

\begin{table*}
\scriptsize

\begin{center}

\begin{tabular}{m{3.5cm}m{3.5cm}m{3.5cm}m{3.5cm}}  
 \toprule
 \textbf{Left context} & \textbf{Ground truth} & \textbf{Retrieved/Generated} & \textbf{Error Analysis} \\ 
 \midrule
  \CXT{ Dog's do not pass a mirror test so it's highly unlikely they have a sense of self.[...]\textbf{They do recognize there names, it's a bit of an illusion though.} } & \EG{ For example if you have two dogs, and you give one of them a treat when you say Fido, and the other when you say Clifford, they learn that the respective words only apply to them.} & [ \hlc[babyblue]{\textit{EGRET}}] For example people can identify their own dog. & \hlc[babyblue]{\textit{EGRET}} retrieved relevant but semantically incorrect example (people identify their dog, instead of how dog identify themselves). \\
  \midrule
  \CXT{An economist would say healthcare has a positive externality.[...]\textbf{There are some things you can buy that make everyone better off.}} & \EG{For example: going to the doctor every time you are sick will make you less likely to make other people sick.} & [\hlc[babyblue]{\textit{EGRET}}] For example, a butterfly house, a free cinema, games consoles etc. & \hlc[babyblue]{\textit{EGRET}} retrieved examples related to the immediate preceding context ("some things you can buy...") but failed to retrieved examples based on earlier context (about healthcare).\\ 
  \midrule
  \CXT{4 billion - Economic and Military aid for Pakistan, Egypt, and Jordan. The goal is to have a few people in the mideast who call us allies. Essentially, we buy their cooperation. \textbf{That cooperation is sometimes useful.}} & \EG{For example, when we killed Osama Bin Laden, we sent troops into Pakistan. Normally, countries don't tolerate troops from other countries. The Pakistanis did complain a little, but they didn't do anything about it.} & [\hlc[brilliantlavender]{\textit{c-REALM-RT}}] For example, for everyone here talking about how a lot of aid works: If we put money towards helping foreign countries rebuild, we are imposing restrictions on domestic activity. [...] & \hlc[brilliantlavender]{\textit{c-REALM-RT}} generated an on-topic hypothetical example, which contradicts with the context. \\ 
  \midrule
  \CXT{You don't usually work on the same files because everything is split up between the departments. \textbf{I haven't used USD yet but I have encountered the following workflow in different studios (using Maya).}} & \EG{For example: a character that has been rigged by one (or more, but not at the same time) rigger goes to the animators. Every animator works with the same character rig BUT each animator works on his/her own shot.[...]} & [\hlc[brilliantlavender]{\textit{c-REALM-RT}}] For example, as a starting point: I'm a post graduate and work in the final sector of the project, not the project itself. Most of the work is done with other studios around the world who are made up of multiple departments.[...] & \hlc[brilliantlavender]{\textit{c-REALM-RT}} generated a personal example that is irrelevant to the context. \\ 
  
 \bottomrule
 
\end{tabular}
\end{center}
\caption{Error analysis of retrieved and generated examples. }
\label{tab:example_error_analysis}

\end{table*}

Our human study is coarse, judging the overall quality of the example, and future experiments could perform more fine-grained ratings of properties such as grammaticality and relevance. 

\section{Related Work}
\label{sec:related_work}




\paragraph{Linguistic studies of exemplification:} 
Early work~\cite{kurohashi-nagao-1994-automatic} studies automatic detection of discourse relations including exemplification. Several works have studied exemplification in the domains of academic writing and teaching~\citep{Hyland2007ApplyingAG, Oliveira2016ExemplificationIS, Triki2021ExemplificationIR}.
They proposed the following categorization of examples: \textit{general example} (an instance of a general category); \textit{analogy} (a parallel or similar case); \textit{counterexample} (example that opposes or contradicts the anchor) and \textit{extreme example} (boundary cases, in the sense of being more of an unusual instance than a representative, generic case). During our initial investigation, we noticed that majority of the examples in our dataset are \textit{general examples}, which aligns with the findings in \citet{Hyland2007ApplyingAG}. The dominance of \textit{general examples} could also be due to the choice of our exemplification markers -- all examples given by \citet{Hyland2007ApplyingAG} for \textit{analogy} have the exemplification marker "like", which we found noisy for automatic extraction.
~\citet{Li2016TheID} examine the closely-related \emph{instantiation} discourse relation, where one text span explains in further detail the events described in another text span.

\paragraph{Long-form question answering:} Our work studies exemplification mainly within the task of long-form question answering (LFQA), which involves generating paragraph-length answers to open-ended questions. Previous work has approached this problem using retrieval-augmented generation~\citep{fan-etal-2019-eli5, Lewis2020RetrievalAugmentedGF}, while~\citet{DBLP:journals/corr/abs-2112-09332} set up an interactive language model that learns LFQA through human interaction.~\citet{lfqa21} demonstrate that lexical overlap metrics such as ROUGE are not meaningful for this task. With a series of human annotations, ~\citet{xu2022lfqadiscourse} study the discourse structures of long-form answers and identify \EG{exemplification} as one of the functional roles commonly present in different types of long-form answers.


\paragraph{Neural retrieval models:} Our \model\ retriever builds on recently-developed neural models that retrieve evidence documents for open-retrieval question answering~\citep{karpukhin-etal-2020-dense, Guu2020REALMRL} and fact-checking~\citep{samarinas-etal-2021-improving}. These models demonstrate superior performance compared to non-neural methods like BM25~\citep{robertson1995okapi}; that said recent sparse/dense hybrid retrievers~\citep{10.1162/tacl_a_00369} could be interesting to explore on our exemplification task in the future.

\section{Conclusion}
\label{sec:conclusion}

In this work, we present the first computational study of exemplification in long-form question answering. We perform a detailed annotation over the use of exemplification across various domains and observe different distributions over complex exemplification types and units. While existing LFQA systems are conditional language models that do not give special treatment to exemplification, we propose to \emph{retrieve} \EG{examples} based on their \CXT{context} instead of generating them. We develop \model, a simple dual encoder trained with a contrastive learning objective, that outperforms a diverse set of baselines on this task of example retrieval, which we can meaningfully evaluate using simple ranking metrics instead of unsuitable metrics like ROUGE. We hope that our work spurs researchers to consider separately modeling and evaluating the fine-grained linguistic and discourse phenomena found in LFQA data.

\section*{Ethical Considerations}

We make use of pretrained language models to both generate and retrieve text in this work. Representations from pretrained language models are known to cause ethical concerns, such as perpetuating racial or gender bias~\citep{field-etal-2021-survey, gender-bias}. We advise using caution and adopting a post-processing strategy to filter potentially offensive text produced by pretrained language models before releasing text content to users. Additionally, we note that most existing LFQA datasets (including the ELI5 dataset used in this work) and benchmarks are collected from English text sources. We hope future works can explore the use of exemplification in other languages.

\section*{Acknowledgements}

We are grateful to the UMass NLP group for many helpful discussions. We thank Yapei Chang, Kalpesh Krishna, Katherine Thai for sharing their retriever models with us, as well as Tu Vu and Andrew Drozdov for sharing their compute resources. We would also like to thank the anonymous reviewers for their insightful feedback. SW and MI were supported by awards IIS-1955567 and IIS-2046248 from the National Science Foundation.

\bibliography{bib/journal-full,bib/anthology,bib/custom}

\begin{thebibliography}{34}
\expandafter\ifx\csname natexlab\endcsname\relax\def\natexlab#1{#1}\fi

\bibitem[{{\"A}del(2006)}]{del2006MetadiscourseIL}
Annelie {\"A}del. 2006.
\newblock Metadiscourse in l1 and l2 english.

\bibitem[{Bajaj et~al.(2018)Bajaj, Campos, Craswell, Deng, Gao, Liu, Majumder,
  McNamara, Mitra, Nguyen, Rosenberg, Song, Stoica, Tiwary, and
  Wang}]{bajaj2018ms}
Payal Bajaj, Daniel Campos, Nick Craswell, Li~Deng, Jianfeng Gao, Xiaodong Liu,
  Rangan Majumder, Andrew McNamara, Bhaskar Mitra, Tri Nguyen, Mir Rosenberg,
  Xia Song, Alina Stoica, Saurabh Tiwary, and Tong Wang. 2018.
\newblock \href {http://arxiv.org/abs/1611.09268} {Ms marco: A human generated
  machine reading comprehension dataset}.

\bibitem[{Chen et~al.(2020)Chen, Kornblith, Norouzi, and
  Hinton}]{chen2020simple}
Ting Chen, Simon Kornblith, Mohammad Norouzi, and Geoffrey Hinton. 2020.
\newblock A simple framework for contrastive learning of visual
  representations.
\newblock In \emph{Proceedings of the International Conference of Machine
  Learning}.

\bibitem[{Clouse(2013)}]{clouse2013student}
Barbara~Fine Clouse. 2013.
\newblock \emph{The student writer}.
\newblock McGraw-Hill.

\bibitem[{Fan et~al.(2019)Fan, Jernite, Perez, Grangier, Weston, and
  Auli}]{fan-etal-2019-eli5}
Angela Fan, Yacine Jernite, Ethan Perez, David Grangier, Jason Weston, and
  Michael Auli. 2019.
\newblock \href {https://doi.org/10.18653/v1/P19-1346} {{ELI}5: Long form
  question answering}.
\newblock In \emph{Proceedings of the 57th Annual Meeting of the Association
  for Computational Linguistics}, pages 3558--3567, Florence, Italy.
  Association for Computational Linguistics.

\bibitem[{Field et~al.(2021)Field, Blodgett, Waseem, and
  Tsvetkov}]{field-etal-2021-survey}
Anjalie Field, Su~Lin Blodgett, Zeerak Waseem, and Yulia Tsvetkov. 2021.
\newblock \href {https://doi.org/10.18653/v1/2021.acl-long.149} {A survey of
  race, racism, and anti-racism in {NLP}}.
\newblock In \emph{Proceedings of the 59th Annual Meeting of the Association
  for Computational Linguistics and the 11th International Joint Conference on
  Natural Language Processing (Volume 1: Long Papers)}, pages 1905--1925,
  Online. Association for Computational Linguistics.

\bibitem[{Gala et~al.(2020)Gala, Khursheed, Lerner, O'Connor, and
  Iyyer}]{gender-bias}
Dhruvil Gala, Mohammad~Omar Khursheed, Hannah Lerner, Brendan O'Connor, and
  Mohit Iyyer. 2020.
\newblock Analyzing gender bias within narrative tropes.
\newblock In \emph{Workshop on NLP and CSS at EMNLP}.

\bibitem[{Gao et~al.(2020)Gao, Biderman, Black, Golding, Hoppe, Foster, Phang,
  He, Thite, Nabeshima, Presser, and Leahy}]{pile}
Leo Gao, Stella Biderman, Sid Black, Laurence Golding, Travis Hoppe, Charles
  Foster, Jason Phang, Horace He, Anish Thite, Noa Nabeshima, Shawn Presser,
  and Connor Leahy. 2020.
\newblock The {P}ile: An 800gb dataset of diverse text for language modeling.
\newblock \emph{arXiv preprint arXiv:2101.00027}.

\bibitem[{Guu et~al.(2020)Guu, Lee, Tung, Pasupat, and Chang}]{Guu2020REALMRL}
Kelvin Guu, Kenton Lee, Zora Tung, Panupong Pasupat, and Ming-Wei Chang. 2020.
\newblock Realm: Retrieval-augmented language model pre-training.
\newblock \emph{ArXiv}, abs/2002.08909.

\bibitem[{Hyland(2007)}]{Hyland2007ApplyingAG}
Ken Hyland. 2007.
\newblock Applying a gloss: exemplifying and reformulating in academic
  discourse.
\newblock \emph{Applied Linguistics}, 28:266--285.

\bibitem[{Karpukhin et~al.(2020)Karpukhin, Oguz, Min, Lewis, Wu, Edunov, Chen,
  and Yih}]{karpukhin-etal-2020-dense}
Vladimir Karpukhin, Barlas Oguz, Sewon Min, Patrick Lewis, Ledell Wu, Sergey
  Edunov, Danqi Chen, and Wen-tau Yih. 2020.
\newblock \href {https://doi.org/10.18653/v1/2020.emnlp-main.550} {Dense
  passage retrieval for open-domain question answering}.
\newblock In \emph{Proceedings of Empirical Methods in Natural Language
  Processing}.

\bibitem[{Khattab and Zaharia(2020)}]{khattab2020colbert}
Omar Khattab and Matei Zaharia. 2020.
\newblock \href {http://arxiv.org/abs/2004.12832} {Colbert: Efficient and
  effective passage search via contextualized late interaction over bert}.

\bibitem[{Kingma and Ba(2015)}]{adam_kingma}
Diederik~P. Kingma and Jimmy Ba. 2015.
\newblock \href {http://arxiv.org/abs/1412.6980} {Adam: {A} method for
  stochastic optimization}.
\newblock In \emph{3rd International Conference on Learning Representations,
  {ICLR} 2015, San Diego, CA, USA, May 7-9, 2015, Conference Track
  Proceedings}.

\bibitem[{Krishna et~al.(2021)Krishna, Roy, and Iyyer}]{lfqa21}
Kalpesh Krishna, Aurko Roy, and Mohit Iyyer. 2021.
\newblock Hurdles to progress in long-form question answering.
\newblock In \emph{North American Association for Computational Linguistics}.

\bibitem[{Kurohashi and Nagao(1994)}]{kurohashi-nagao-1994-automatic}
Sadao Kurohashi and Makoto Nagao. 1994.
\newblock \href {https://aclanthology.org/C94-2183} {Automatic detection of
  discourse structure by checking surface information in sentences}.
\newblock In \emph{{COLING} 1994 Volume 2: The 15th {I}nternational
  {C}onference on {C}omputational {L}inguistics}.

\bibitem[{Kwiatkowski et~al.(2019)Kwiatkowski, Palomaki, Redfield, Collins,
  Parikh, Alberti, Epstein, Polosukhin, Kelcey, Devlin, Lee, Toutanova, Jones,
  Chang, Dai, Uszkoreit, Le, and Petrov}]{natural_questions}
Tom Kwiatkowski, Jennimaria Palomaki, Olivia Redfield, Michael Collins, Ankur
  Parikh, Chris Alberti, Danielle Epstein, Illia Polosukhin, Matthew Kelcey,
  Jacob Devlin, Kenton Lee, Kristina~N. Toutanova, Llion Jones, Ming-Wei Chang,
  Andrew Dai, Jakob Uszkoreit, Quoc Le, and Slav Petrov. 2019.
\newblock Natural questions: a benchmark for question answering research.
\newblock \emph{Transactions of the Association of Computational Linguistics}.

\bibitem[{Lewis et~al.(2020)Lewis, Perez, Piktus, Petroni, Karpukhin, Goyal,
  Kuttler, Lewis, tau Yih, Rockt{\"a}schel, Riedel, and
  Kiela}]{Lewis2020RetrievalAugmentedGF}
Patrick Lewis, Ethan Perez, Aleksandara Piktus, Fabio Petroni, Vladimir
  Karpukhin, Naman Goyal, Heinrich Kuttler, Mike Lewis, Wen tau Yih, Tim
  Rockt{\"a}schel, Sebastian Riedel, and Douwe Kiela. 2020.
\newblock Retrieval-augmented generation for knowledge-intensive nlp tasks.
\newblock \emph{ArXiv}, abs/2005.11401.

\bibitem[{Li and Nenkova(2016)}]{Li2016TheID}
Junyi~Jessy Li and Ani Nenkova. 2016.
\newblock The instantiation discourse relation: A corpus analysis of its
  properties and improved detection.
\newblock In \emph{NAACL}.

\bibitem[{Lin(2004)}]{lin2004rouge}
Chin-Yew Lin. 2004.
\newblock Rouge: A package for automatic evaluation of summaries.
\newblock In \emph{Text summarization branches out}, pages 74--81.

\bibitem[{Liu et~al.(2019)Liu, Ott, Goyal, Du, Joshi, Chen, Levy, Lewis,
  Zettlemoyer, and Stoyanov}]{liu2019roberta}
Yinhan Liu, Myle Ott, Naman Goyal, Jingfei Du, Mandar Joshi, Danqi Chen, Omer
  Levy, Mike Lewis, Luke Zettlemoyer, and Veselin Stoyanov. 2019.
\newblock Roberta: A robustly optimized {BERT} pretraining approach.
\newblock \emph{arXiv preprint arXiv:1907.11692}.

\bibitem[{Luan et~al.(2021)Luan, Eisenstein, Toutanova, and
  Collins}]{10.1162/tacl_a_00369}
Yi~Luan, Jacob Eisenstein, Kristina Toutanova, and Michael Collins. 2021.
\newblock \href {https://doi.org/10.1162/tacl_a_00369} {{Sparse, Dense, and
  Attentional Representations for Text Retrieval}}.
\newblock \emph{Transactions of the Association for Computational Linguistics},
  9:329--345.

\bibitem[{Meyer(1992)}]{Meyer1992AppositionIC}
Charles~F. Meyer. 1992.
\newblock Apposition in contemporary english.

\bibitem[{Nakano et~al.(2021)Nakano, Hilton, Balaji, Wu, Ouyang, Kim, Hesse,
  Jain, Kosaraju, Saunders, Jiang, Cobbe, Eloundou, Krueger, Button, Knight,
  Chess, and Schulman}]{DBLP:journals/corr/abs-2112-09332}
Reiichiro Nakano, Jacob Hilton, Suchir Balaji, Jeff Wu, Long Ouyang, Christina
  Kim, Christopher Hesse, Shantanu Jain, Vineet Kosaraju, William Saunders,
  Xu~Jiang, Karl Cobbe, Tyna Eloundou, Gretchen Krueger, Kevin Button, Matthew
  Knight, Benjamin Chess, and John Schulman. 2021.
\newblock \href {http://arxiv.org/abs/2112.09332} {Webgpt: Browser-assisted
  question-answering with human feedback}.
\newblock \emph{CoRR}, abs/2112.09332.

\bibitem[{Oliveira and Brown(2016)}]{Oliveira2016ExemplificationIS}
Alandeom~W. Oliveira and Adam~Oliver Brown. 2016.
\newblock Exemplification in science instruction: Teaching and learning through
  examples.
\newblock \emph{Journal of Research in Science Teaching}, 53:737--767.

\bibitem[{Petroni et~al.(2021)Petroni, Piktus, Fan, Lewis, Yazdani, De~Cao,
  Thorne, Jernite, Karpukhin, Maillard, Plachouras, Rockt{\"a}schel, and
  Riedel}]{petroni-etal-2021-kilt}
Fabio Petroni, Aleksandra Piktus, Angela Fan, Patrick Lewis, Majid Yazdani,
  Nicola De~Cao, James Thorne, Yacine Jernite, Vladimir Karpukhin, Jean
  Maillard, Vassilis Plachouras, Tim Rockt{\"a}schel, and Sebastian Riedel.
  2021.
\newblock \href {https://doi.org/10.18653/v1/2021.naacl-main.200} {{KILT}: a
  benchmark for knowledge intensive language tasks}.
\newblock In \emph{Proceedings of the 2021 Conference of the North American
  Chapter of the Association for Computational Linguistics: Human Language
  Technologies}, pages 2523--2544, Online. Association for Computational
  Linguistics.

\bibitem[{Rae et~al.(2020)Rae, Potapenko, Jayakumar, Hillier, and
  Lillicrap}]{Rae2020Compressive}
Jack~W. Rae, Anna Potapenko, Siddhant~M. Jayakumar, Chloe Hillier, and
  Timothy~P. Lillicrap. 2020.
\newblock \href {https://openreview.net/forum?id=SylKikSYDH} {Compressive
  transformers for long-range sequence modelling}.
\newblock In \emph{International Conference on Learning Representations}.

\bibitem[{Reimers and Gurevych(2021)}]{reimers-2020-Curse_Dense_Retrieval}
Nils Reimers and Iryna Gurevych. 2021.
\newblock \href {https://arxiv.org/abs/2012.14210} {The curse of dense
  low-dimensional information retrieval for large index sizes}.
\newblock In \emph{Proceedings of the 59th Annual Meeting of the Association
  for Computational Linguistics and the 11th International Joint Conference on
  Natural Language Processing (Volume 2: Short Papers)}, pages 605--611,
  Online. Association for Computational Linguistics.

\bibitem[{Robertson et~al.(1995)Robertson, Walker, Jones, Hancock-Beaulieu,
  Gatford et~al.}]{robertson1995okapi}
Stephen~E Robertson, Steve Walker, Susan Jones, Micheline~M Hancock-Beaulieu,
  Mike Gatford, et~al. 1995.
\newblock Okapi at trec-3.
\newblock \emph{Nist Special Publication Sp}, 109:109.

\bibitem[{Samarinas et~al.(2021)Samarinas, Hsu, and
  Lee}]{samarinas-etal-2021-improving}
Chris Samarinas, Wynne Hsu, and Mong~Li Lee. 2021.
\newblock \href {https://doi.org/10.18653/v1/2021.naacl-demos.10} {Improving
  evidence retrieval for automated explainable fact-checking}.
\newblock In \emph{Proceedings of the 2021 Conference of the North American
  Chapter of the Association for Computational Linguistics: Human Language
  Technologies: Demonstrations}, pages 84--91, Online. Association for
  Computational Linguistics.

\bibitem[{Thai et~al.(2022)Thai, Chang, Krishna, and Iyyer}]{thairelic22}
Katherine Thai, Yapei Chang, Kalpesh Krishna, and Mohit Iyyer. 2022.
\newblock Relic: Retrieving evidence for literary claims.
\newblock In \emph{Association of Computational Linguistics}.

\bibitem[{Triki(2021)}]{Triki2021ExemplificationIR}
Nesrine Triki. 2021.
\newblock Exemplification in research articles: Structural, semantic and
  metadiscursive properties across disciplines.
\newblock \emph{Journal of English for Academic Purposes}, 54:101039.

\bibitem[{van~den Oord et~al.(2018)van~den Oord, Li, and Vinyals}]{infoNCE}
A{\"a}ron van~den Oord, Yazhe Li, and Oriol Vinyals. 2018.
\newblock Representation learning with contrastive predictive coding.
\newblock \emph{ArXiv}, abs/1807.03748.

\bibitem[{Vaswani et~al.(2017)Vaswani, Shazeer, Parmar, Uszkoreit, Jones,
  Gomez, Kaiser, and Polosukhin}]{Vaswani2017AttentionIA}
Ashish Vaswani, Noam~M. Shazeer, Niki Parmar, Jakob Uszkoreit, Llion Jones,
  Aidan~N. Gomez, Lukasz Kaiser, and Illia Polosukhin. 2017.
\newblock Attention is all you need.
\newblock \emph{ArXiv}, abs/1706.03762.

\bibitem[{Xu et~al.(2022)Xu, Li, and Choi}]{xu2022lfqadiscourse}
Fangyuan Xu, Junyi~Jessy Li, and Eunsol Choi. 2022.
\newblock How do we answer complex questions: Discourse structure of long-form
  answers.
\newblock In \emph{Proceedings of the Annual Meeting of the Association for
  Computational Linguistics}.
\newblock Long paper.

\end{thebibliography}
\bibliographystyle{bib/acl_natbib}

\clearpage
\setcounter{table}{0}
\renewcommand{\thetable}{A\arabic{table}}
\appendix





\section{Evaluation Interface on Mechanic Turk}
 Given a question, a partial answer (context) and three candidate examples, a worker is asked to rate all three candidate examples, according to their fit with the question and the given context.

\begin{tabular}{c}
 \includegraphics[align=c,scale=0.26]{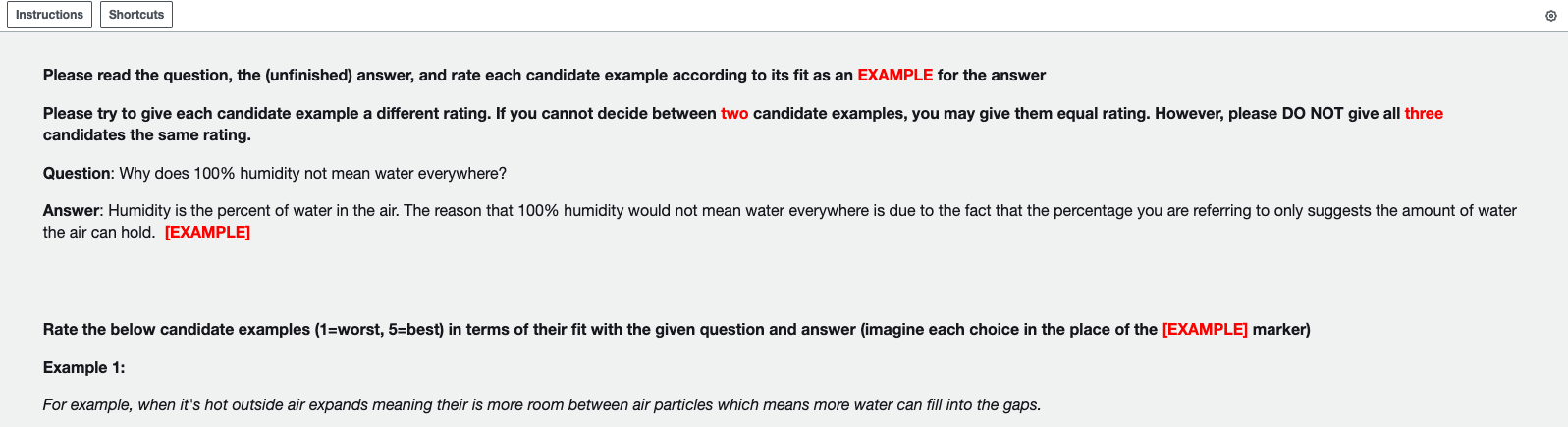} \\
 \includegraphics[align=c,scale=0.26]{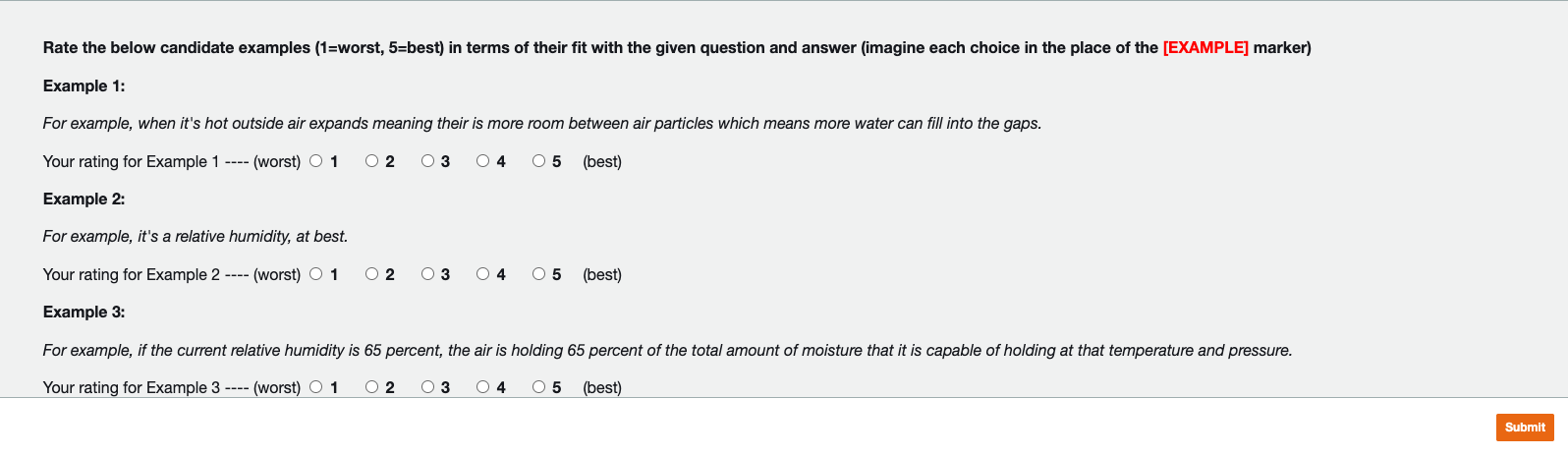} \\
\end{tabular}

\clearpage

\section{Variations in Recall@k}
\label{appendix:recall@k}
\begin{figure}[h]
    \begin{minipage}{0.45\textwidth}
        \centering
        \includegraphics[width=1.1\textwidth]{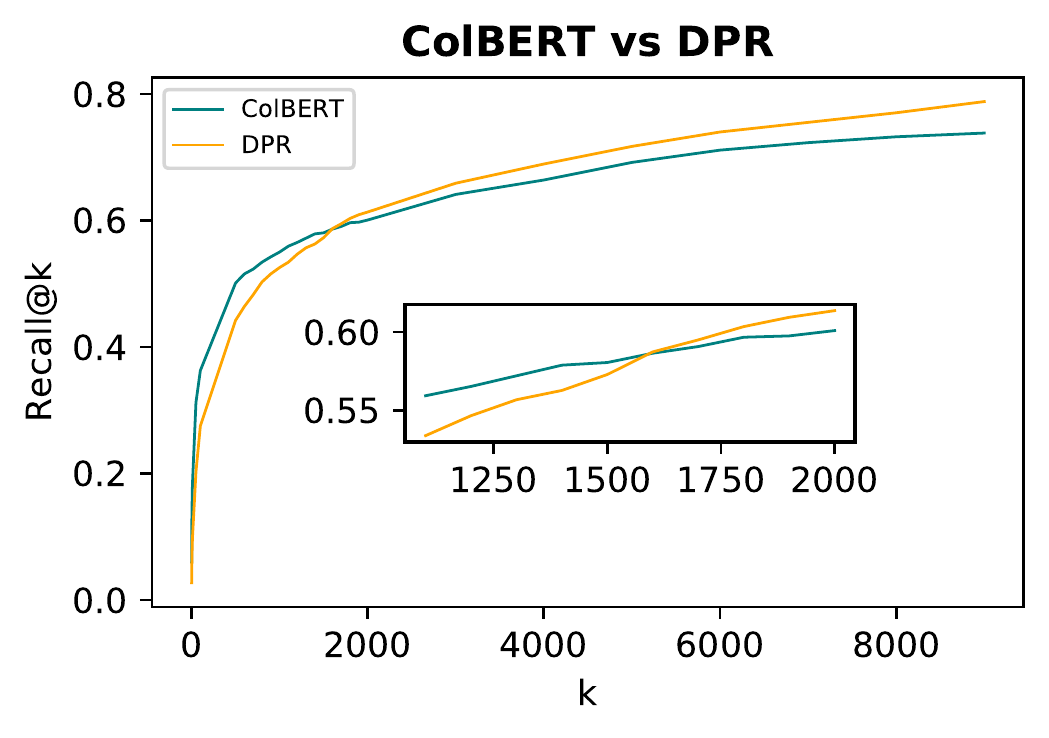} 
        \caption{ColBERT gives higher recall@k compared to DPR when $k$ is relatively small but lower recall@k when $k$ is larger}
        \label{fig:colbert-dpr}

    \end{minipage}\hfill
    \begin{minipage}{0.45\textwidth}
        \centering
        \includegraphics[width=1.1\textwidth]{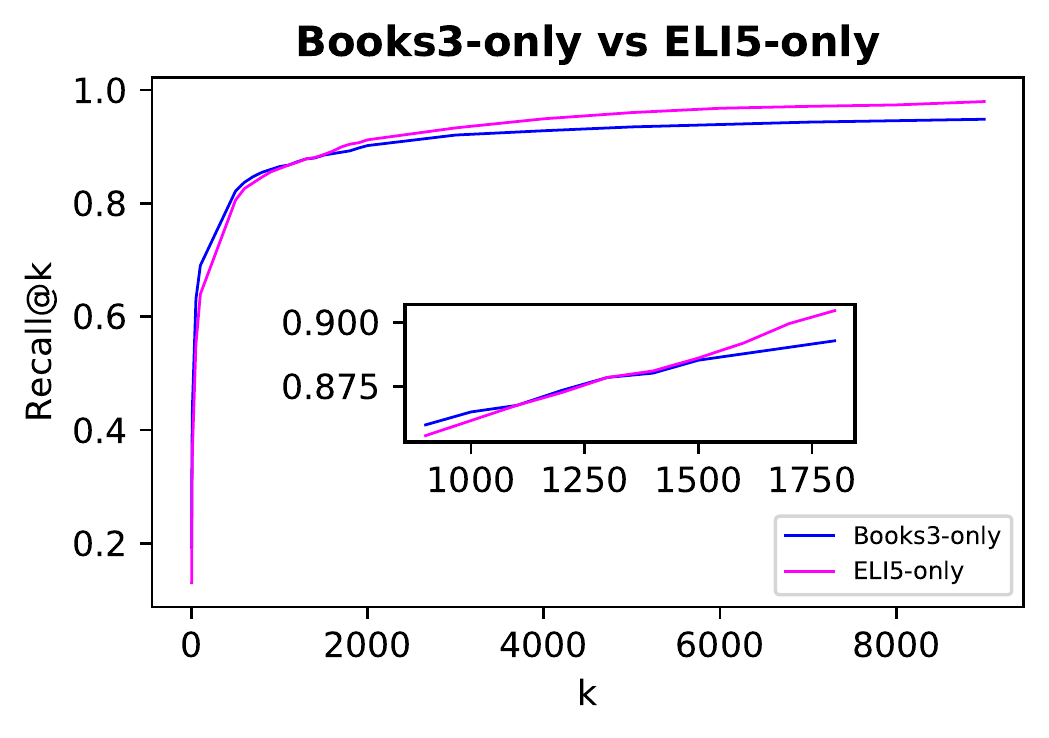} 
        \caption{\model-Books3-only gives higher recall@k compared to \model-ELI5-only when $k$ is relatively small but lower recall@k when $k$ is larger}    
        \label{fig:books3-eli5}
    \end{minipage}
\end{figure}

Table \ref{tab:retriever_stats} shows that ColBERT outperforms DPR in various recall@k measures when $k$ is relatively small (less than 100) and yet underperforms DPR in the average rank. Similar phenomenon occurs with \model-Books3-only and \model-ELI5-only too. We further computed the recall@k for all these models when $k$ is very large (close to $10,000$). Fig. \ref{fig:colbert-dpr} and Fig. \ref{fig:books3-eli5} show that despite their lower recall@k when $k$ is relatively small, DPR and \model-ELI5-only yield higher recall@k when $k$ is relatively large, compared to ColBERT and \model-Books3-only respectively. With better performance at the long tail (when $k$ is relatively large),  DPR and \model-ELI5-only also produce lower average ranks in Table \ref{tab:retriever_stats}.

\begin{table*}
\small

\begin{center}

\begin{tabular}{m{0.8cm}m{1.5cm}m{1cm}m{5cm}m{5cm}}  
 \toprule
 \textbf{Dataset} & \textbf{Type} & \textbf{Personal} & \textbf{Left context} & \textbf{Extracted Example} \\ 
  \midrule
  NQ & Real && \CXT{Group Areas Act was the title of three acts of the Parliament of South Africa enacted under the apartheid government of South Africa. The acts assigned racial groups to different residential and business sections in urban areas in a system of urban apartheid.\textbf{An effect of the law was to exclude non-Whites from living in the most developed areas , which were restricted to Whites} } & ( e.g. , \EG{\textit{Sea Point , Lansdowne , Cape Town , Claremont , Cape Town })}. \\
  \midrule
  NQ & Hypothetical && \CXT{Although the safest way to recognize a chord 's root is , after having reduced the chord to close spacing , to rearrange it as a stack of thirds , there are shortcuts to this : [...] \textbf{With chord types, such as chords with added sixths or chords over pedal points, more than one possible chordal analysis may be possible.}} & For example, \EG{\textit{in a tonal piece of music , the notes C, E, G, A, sounded as a chord , could be analyzed as a C major sixth chord in root position ( a major triad -- C, E, G -- with an added sixth -- A -- above the root ) or as a first inversion A minor seventh chord ( the A minor seventh chord contains the notes A, C, E and G, but in this example, the C note, the third of the A minor chord, is in the bass ).}}\\
  \midrule
  ELI5 & Real & \checkmark & \CXT{my uncle owns a pretty large recycling business. They export the majority of their newly created raw materials to the places that produce with the materials (China).[...] \textbf{Raw material are often remade into base products several times over before it gets to a manufacturing plant.}} & For example: \EG{\textit{My uncle’s business is primarily plastics. They get cast offs, seconds, etc plastic from all kinds of US manufacturers. They then sort, filter and break down the plastic to the most basic starting point (often really small non died beads) and ship it to China. [...]}}\\ 
  \midrule
  ELI5 & Hypothetical & \checkmark & \CXT{OP I guess you are coming from movies/ace attorney but avoiding that. Let's say you have the most cut and dry murder case [...] \textbf{There are a limited number of prosecutors, judges, and defense attorneys}} & (for example \EG{\textit{I currently intern at a medical malpractice firm and if we were forced to do criminal defense I would actually be the most qualified one there to do so- at a firm where the youngest attorney still has 15 years of experience}})\\ 
  \midrule
  Books3 & Real && \CXT{\textbf{People in a second group were given a verbal description, with which they were to construct an image of walking along the two segments}} & For example, \EG{\textit{people were told to imagine they would "Go forward 3 m, turn clockwise 90°, then go forward 3 m."}} \\ \midrule 
  Books3 & Hypothetical && \CXT{When we cook together, I have to stay alert because she is always throwing a lemon at me—\textbf{sometimes double down on acid and mix lemon juice with a little bit of vinegar to get the sunny sweet-sour note of the citrus along with earthy, apple, or wine notes of a vinegar for greater complexity}} & For example, \EG{\textit{ if you toss roasted beets (a notoriously earthy and sweet vegetable that some might say tastes like soil) with just lemon juice, olive oil, and salt, it would no doubt be good, but if you supplement the sunny lemon juice with a tiny splash of sherry vinegar for its woodsy earthiness, you get a roasted beet dish that is far more complex and delicious than if you had used only one or the other.}} \\ 
  
 
 \bottomrule
 
\end{tabular}

\end{center}
\caption{Different types of annotated \EG{examples} in the three datasets. The \CXT{\textbf{anchor}} and \EG{\textit{example}} are highlighted.}
\label{tab:annotated_example}

\end{table*}

\end{document}